\def\year{2020}\relax
\documentclass[letterpaper]{article} 
\usepackage{aaai20}  
\usepackage{times}  
\usepackage{helvet} 
\usepackage{courier}  
\usepackage[hyphens]{url}  
\usepackage{graphicx} 
\usepackage{graphicx}  
\usepackage{amssymb}
\usepackage{times}
\usepackage{latexsym}
\usepackage{graphicx}  
\usepackage{subfigure}
\usepackage{url}
\usepackage{amsfonts} 
\usepackage{amsmath}
\usepackage{booktabs}
\usepackage{multirow}
\usepackage{arydshln}

\title{Semantics-aware BERT for Language Understanding}
\author{
	Zhuosheng Zhang\textsuperscript{\rm 1,2,3,\thanks{These authors contribute equally. $\dagger$Corresponding author. This paper was partially supported by National Key Research and Development Program of China (No. 2017YFB0304100) and Key Projects of National Natural Science Foundation of China (U1836222 and 61733011).}},
	Yuwei Wu\textsuperscript{\rm 1,2,3,4,*},
	Hai Zhao\textsuperscript{\rm 1,2,3,$\dagger$},
	Zuchao Li\textsuperscript{\rm 1,2,3}, \\
	\large \textbf{ 
		Shuailiang Zhang\textsuperscript{\rm 1,2,3},
		Xi Zhou\textsuperscript{\rm 5},
		Xiang Zhou\textsuperscript{\rm 5}
	}
	\\
	\textsuperscript{\rm 1}Department of Computer Science and Engineering, Shanghai Jiao Tong University\\
	\textsuperscript{\rm 2}Key Laboratory of Shanghai Education Commission for Intelligent Interaction\\
	and Cognitive Engineering, Shanghai Jiao Tong University, Shanghai, China\\
	\textsuperscript{\rm 3}MoE Key Lab of Artificial Intelligence, AI Institute, Shanghai Jiao Tong University, Shanghai, China\\
	\textsuperscript{\rm 4}College of Zhiyuan, Shanghai Jiao Tong University, China\\
	\textsuperscript{\rm 5}CloudWalk Technology, Shanghai, China\\
	{\tt \{zhangzs,will8821\}@sjtu.edu.cn, zhaohai@cs.sjtu.edu.cn} \\
}
 
\begin{document}

\maketitle

\begin{abstract}

	The latest work on language representations carefully integrates contextualized features into language model training, which enables a series of success especially in various machine reading comprehension and natural language inference tasks. However, the existing language representation models including ELMo, GPT and BERT only exploit plain context-sensitive features such as character or word embeddings. They rarely consider incorporating structured semantic information which can provide rich semantics for language representation. To promote natural language understanding, we propose to incorporate explicit contextual semantics from pre-trained semantic role labeling, and introduce an improved language representation model, Semantics-aware BERT (SemBERT), which is capable of explicitly absorbing contextual semantics over a BERT backbone. SemBERT keeps the convenient usability of its BERT precursor in a light fine-tuning way without substantial task-specific modifications. Compared with BERT, semantics-aware BERT is as simple in concept but more powerful. It obtains new state-of-the-art or substantially improves results on ten reading comprehension and language inference tasks. 

\end{abstract}

\section{Introduction}

Recently, deep contextual language model (LM) has been shown effective for learning universal language representations, achieving state-of-the-art results in a series of flagship natural language understanding (NLU) tasks. Some prominent examples are Embedding from Language models (ELMo) \cite{Peters2018ELMO}, Generative Pre-trained Transformer (OpenAI GPT)  \cite{radford2018improving}, Bidirectional Encoder Representations from Transformers (BERT) \cite{devlin2018bert} and Generalized Autoregressive Pretraining (XLNet) \cite{yang2019xlnet}. Providing fine-grained contextual embedding, these pre-trained models could be either easily applied to downstream models as the encoder or used for fine-tuning. 

Despite the success of those well pre-trained language models, we argue that current techniques which only focus on language modeling restrict the power of the pre-trained representations. The major limitation of existing language models lies in only taking plain contextual features for both representation and training objective, rarely considering explicit contextual semantic clues. Even though well pre-trained language models can implicitly represent contextual semantics more or less \cite{clark2019does}, they can be further enhanced by incorporating external knowledge. To this end, there is a recent trend of incorporating extra knowledge to pre-trained language models \cite{zhang2019sg}.

A number of studies have found deep learning models might not really understand the natural language texts \cite{Mudrakarta2018Did} and vulnerably suffer from adversarial attacks \cite{Jia2017Adversarial}. Through their observation, deep learning models pay great attention to non-significant words and ignore important ones. For attractive question answering challenge \cite{Rajpurkar2016SQuAD}, we observe a number of answers produced by previous models are semantically incomplete (As shown in Section \ref{span_out}), which suggests that the current NLU models suffer from insufficient contextual semantic representation and learning.

Actually, NLU tasks share the similar task purpose as sentence contextual semantic analysis. Briefly,  
semantic role labeling (SRL) over a sentence is to discover \emph{who did what to whom}, \emph{when and why} with respect to the central meaning of the sentence, which naturally matches the task target of NLU. For example, in question answering tasks, questions are usually formed with who, what, how, when and why, which can be conveniently formulized into the predicate-argument relationship in terms of contextual semantics.

In human language, a sentence usually involves various predicate-argument structures, while neural models encode sentence into embedding representation, with little consideration of the modeling of multiple semantic structures. Thus we are motivated to enrich the sentence contextual semantics in multiple predicate-specific argument sequences by presenting SemBERT: Semantics-aware BERT, which is a fine-tuned BERT with explicit contextual semantic clues. The proposed SemBERT learns the representation in a fine-grained manner and takes both strengths of BERT on plain context representation and explicit semantics for deeper meaning representation.

Our model consists of three components:  1) an out-of-shelf semantic role labeler to annotate the input sentences with a variety of semantic role labels; 2) an sequence encoder where a pre-trained language model is  used  to  build  representation  for input raw texts and the semantic role labels are mapped to embedding in parallel;  3) a semantic integration component to integrate the text representation with the contextual explicit semantic embedding to obtain the joint representation for downstream tasks. 

The proposed SemBERT will be directly applied to typical NLU tasks. Our model is evaluated on 11 benchmark datasets involving natural language inference, question answering, semantic similarity and text classification. SemBERT obtains new state-of-the-art on SNLI and also obtains significant gains on the GLUE benchmark and SQuAD 2.0. Ablation studies and analysis verify that our introduced explicit semantics is essential to the further performance improvement and SemBERT essentially and effectively works as a unified semantics-enriched language representation model\footnote{The code is publicly available at \url{https://github.com/cooelf/SemBERT}.}.

\section{Background and Related Work}\label{sec:related}
\subsection{Language Modeling for NLU}
Natural language understanding tasks require a comprehensive understanding of natural languages and the ability to do further inference and reasoning. A common trend among NLU studies is that models are becoming more and more sophisticated with stacked attention mechanisms or large amount of corpus \cite{zhang2018DUA,zhang2019dual,zhou2019limit}, resulting in explosive growth of computational cost. Notably, well pre-trained contextual language models such as ELMo \cite{Peters2018ELMO}, GPT \cite{radford2018improving} and BERT \cite{devlin2018bert} have been shown powerful to boost NLU tasks to reach new high performance. 

Distributed representations have been widely used as a standard part of NLP models due to the ability to capture the local co-occurence of words from large scale unlabeled text \cite{mikolov2013distributed}. However, these approaches for learning word vectors only involve a single, context independent representation for each word with litter consideration of contextual encoding in sentence level. Thus recently introduced contextual language models including ELMo, GPT, BERT and XLNet fill the gap by strengthening the contextual sentence modeling for better representation, among which BERT uses a different pre-training objective, masked language model, which allows capturing both sides of context, left and right. Besides, BERT also introduces a \emph{next sentence prediction} task that jointly pre-trains text-pair representations. The latest evaluation shows that BERT is powerful and convenient for downstream NLU tasks.

The major technical improvement over traditional embeddings of these newly proposed language models is that they focus on extracting context-sensitive features from language models. When integrating these contextual word embeddings with existing task-specific architectures, ELMo helps boost several major NLP benchmarks \cite{Peters2018ELMO} including question answering on SQuAD, sentiment analysis \cite{socher2013recursive}, and named entity recognition \cite{sang2003introduction}, while BERT especially shows effective on language understanding tasks on GLUE, MultiNLI and SQuAD \cite{devlin2018bert}. In this work, we follow this line of extracting context-sensitive features and take pre-trained BERT as our backbone encoder for jointly learning explicit context semantics.

\subsection{Explicit Contextual Semantics}
Although distributed representations including the latest advanced pre-trained contextual language models have already been strengthened by semantics to some extent from linguistic sense \cite{clark2019does}, we argue such implicit semantics may not be enough to support a powerful contextual representation for NLU, according to our observation on the semantically incomplete answer span generated by BERT on SQuAD, which motivates us to directly introduce explicit semantics.

There are a few formal semantic frames, including FrameNet \cite{baker1998berkeley} and PropBank \cite{palmer2005proposition}, in which the latter is more popularly implemented in computational linguistics. Formal semantics generally presents the semantic relationship as predicate-argument structure. For example, given the following sentence with target verb (predicate) \emph{sold}, all the arguments are labeled as follows, 

[$_{ARG0}$ Charlie] [$_{V}$ sold] [$_{ARG1}$ a book] [$_{ARG2}$ to Sherry] [$_{AM-TMP}$ last week].

\noindent where $ARG0$ represents the  seller (agent), $ARG1$ represents the thing sold (theme), $ARG2$ represents the buyer (recipient), $AM-TMP$ is an adjunct indicating the timing of the action and $V$ represents the predicate.

\begin{figure*}[!htb]
	\centering
	\includegraphics[width=0.85\textwidth]{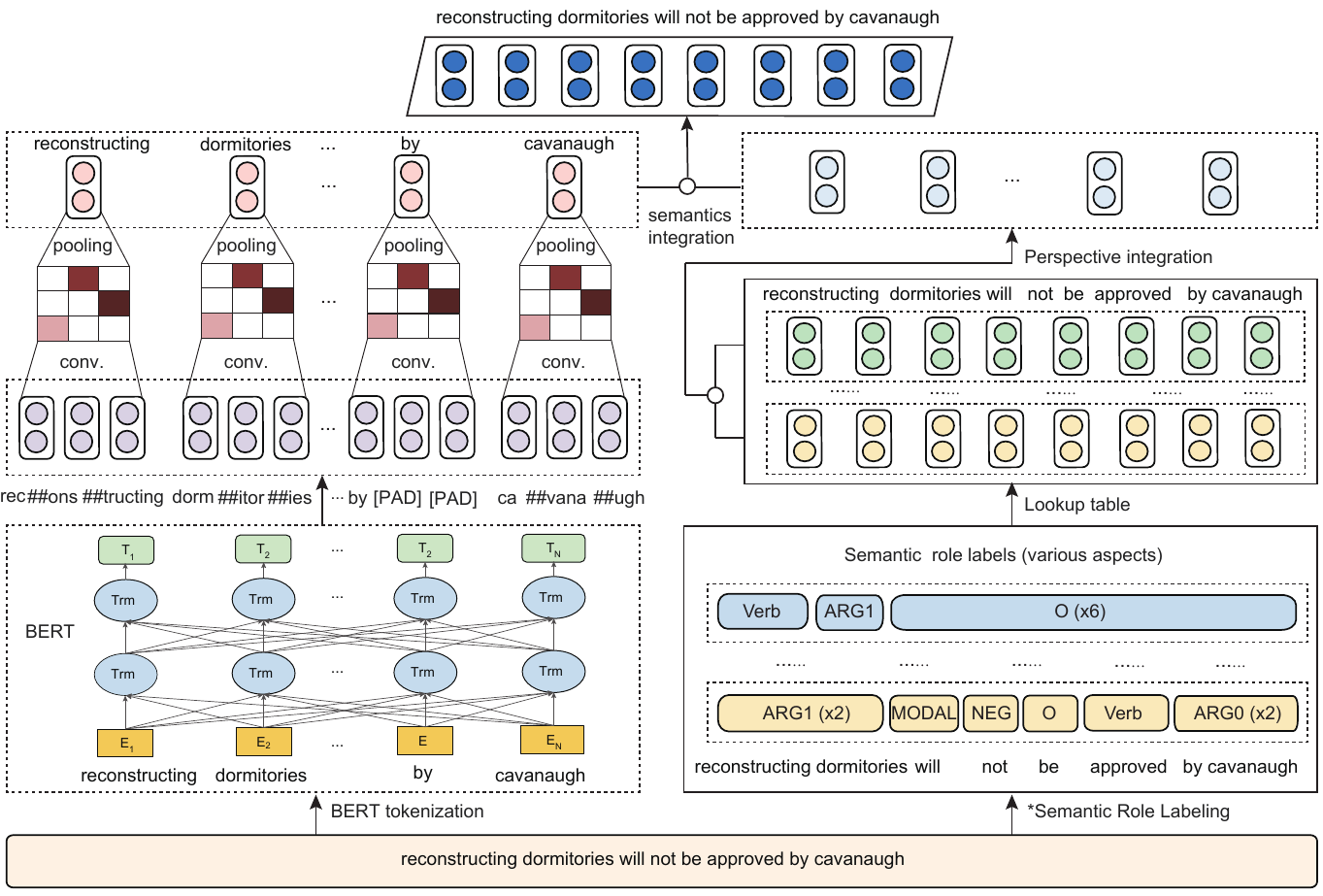}
	\\\begin{flushleft}
		\scriptsize{For the text, \{\emph{reconstructing dormitories will not be approved by cavanaugh}\}, it will be tokenized to a subword-level sequence, \{\emph{rec, \#\#ons, \#\#tructing, dorm, \#\#itor, \#\#ies, will, not, be, approved, by, ca, \#\#vana, \#\#ugh}\}. Meanwhile, there are two kinds of word-level semantic structures,
			
[ARG1: reconstructing dormitories] [ARGM-MOD: will] [ARGM-NEG: not] be [V: approved] [ARG0: by cavanaugh]
	
[V: reconstructing] [ARG1: dormitories] will not be approved by cavanaugh}
	\end{flushleft}
	\caption{\label{fig:framework}Semantics-aware BERT. * denotes the pre-trained labeler which will not be fine-tuned in our framework.}
	
\end{figure*}
To parse the predicate-argument structure, we have an NLP task, semantic role labeling (SRL) \cite{zhao2009semantic,zhao2013integrative}. Recently, end-to-end SRL system neural models have been introduced \cite{He2017Deep,li2019dependency}. These studies tackle argument identification and argument classification in one shot.  \citeauthor{He2017Deep} (\citeyear{He2017Deep}) presented a deep highway BiLSTM architecture with constrained decoding, which is simple and effective, enabling us to select it as our basic semantic role labeler. Inspired by recent advances, we can easily integrate SRL into NLU.

\section{Semantics-aware BERT}\label{sec:bert}
Figure \ref{fig:framework} overviews our semantics-aware BERT framework. We omit rather extensive formulations of BERT and recommend readers to get the details from \cite{devlin2018bert}. SemBERT is designed to be capable of handling multiple sequence inputs. In SemBERT, words in the input sequence are passed to semantic role labeler to fetch multiple predicate-derived structures of explicit semantics and the corresponding embeddings are aggregated after a linear layer to form the final semantic embedding. In parallel, the input sequence is segmented to subwords (if any) by BERT word-piece tokenizer, then the subword representation is transformed back to word level via a convolutional layer to obtain the contextual word representations. At last, the word representations and semantic embedding are concatenated to form the joint representation for downstream tasks.

\begin{figure*}
	\centering
	\includegraphics[width=1.0\textwidth]{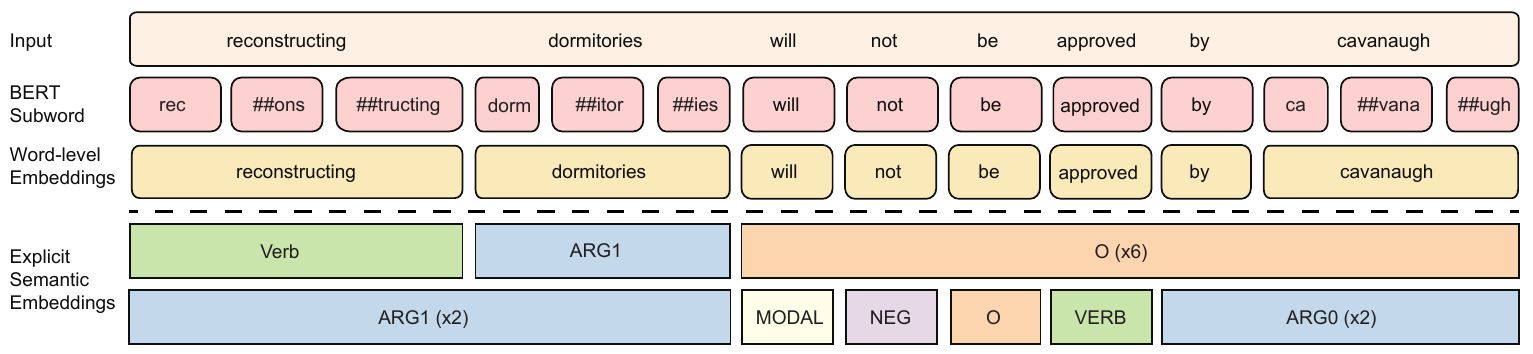}
	\caption{\label{fig:srl_exp}The input representation flow.}
\end{figure*}

\subsection{Semantic Role Labeling}
During the data pre-processing, each sentence is annotated into several semantic sequences using our pre-trained semantic labeler. We take PropBank style \cite{palmer2005proposition} of semantic roles to annotate every token of input sequence with semantic labels. Given a specific sentence, there would be various predicate-argument structures. As shown in Figure \ref{fig:framework}, for the text, [\emph{reconstructing dormitories will not be approved by cavanaugh}], there are two semantic structures in the view of the predicates in the sentence.

To disclose the multidimensional semantics, we group the semantic labels and integrate them with text embeddings in the next encoding component. The input data flow is depicted in Figure \ref{fig:srl_exp}.

\subsection{Encoding}
The raw text sequences and semantic role label sequences are firstly represented as embedding vectors to feed a pre-trained BERT. The input sentence $X=\{x_1, \dots, x_n\}$ is a sequence of words of length $n$, which is first tokenized to word pieces (subword tokens). Then the transformer encoder captures the contextual information for each token via self-attention and produces a sequence of contextual embeddings. 

For $m$ label sequences related to each predicate, we have $T=\{t_1, \dots, t_m\}$ where $t_i$ contains $n$ labels denoted as $\{label^i_1, label^i_2, ..., label^i_n\}$. Since our labels are in word-level, the length is equal to the original sentence length $n$ of $X$. We regard the semantic signals as embeddings and use a lookup table to map these labels to vectors $\{v^i_1, v^i_2, ..., v^i_n\}$ and feed a BiGRU layer to obtain the label representations for $m$ label sequences in latent space, $e(t_i)=BiGRU(v^i_1, v^i_2, \dots, v^i_n)$ where $0< i \leqslant  m$. For $m$ label sequences, let $L_i$ denote the label sequences for token $x_i$, we have $e(L_i)=\{e(t_1), \dots, e(t_m)\}$. We concatenate the $m$ sequences of label representation and feed them to a fully connected layer to obtain the refined joint representation $e^t$ in dimension $d$:
\begin{equation}
\begin{split}
e'(L_i) &= W_2\left[e(t_1), e(t_2), \dots, e(t_{m})\right] + b_2, \\
e^t &= \{e'(L_1),...,e'(L_n)\},
\end{split}
\end{equation}
where $W_2$ and $b_2$ are trainable parameters.

\subsection{Integration} 

This integration module fuses the lexical text embedding and label representations. As the original pre-trained BERT is based on a sequence of subwords, while our introduced semantic labels are on words, we need to align these different sized sequences. Thus we group the subwords for each word and use convolutional neural network (CNN) with a max pooling to obtain the representation in word-level. We select CNN because of fast speed and our preliminary experiments show that it also gives better results than RNNs in our concerned tasks where we think the local feature captured by CNN would be beneficial for subword-derived LM modeling. 

We take one word for example. Supposing that word $x_i$ is made up of a sequence of subwords $[s_1, s_2, ..., s_l]$, where $l$ is the number of subwords for word $x_i$. Denoting the representation of subword $s_j$ from BERT as $e(s_{j})$, we first utilize a Conv1D layer, $e{'_i} = W_1\left[e(s_i), e(s_{i+1}), \dots, e(s_{i+k-1})\right] + b_1$,
where $W_1$ and $b_1$ are trainable parameters and $k$ is the kernel size. We then apply ReLU and max pooling to the output embedding sequence for $x_i$:
\begin{equation}
e_i^* = ReLU(e_i'), e(x_i) = MaxPooling({e_1^*,...,e_{l-k+1}^*}),
\end{equation}
Therefore, the whole representation for word sequence $X$ is represented as $e^w =\{e(x_1), \dots e(x_n)\} \in \mathbb{R}^{n \times d_w}$ where $d_w$ denotes the dimension of word embedding.

The aligned context and distilled semantic embeddings are then merged by a fusion function $h = e^w \diamond e^t$, where $\diamond$ represents concatenation operation\footnote{We also tried summation, multiplication and attention mechanisms, but our experiments show that concatenation is the best.}.

\section{Model Implementation}\label{sec:imp}
Now, we introduce the specific implementation parts of our SemBERT. SemBERT could be a forepart encoder for a wide range of tasks and could also become an end-to-end model with only a linear layer for prediction. For simplicity, we only show the straightforward SemBERT that directly gives the predictions after fine-tuning\footnote{We only use \emph{single} model for each task without jointly training and parameter sharing.}. 

\subsection{Semantic Role Labeler}\label{srl}
To obtain the semantic labels, we use a pre-trained SRL module to predict all predicates and corresponding arguments in one shot. We implement the semantic role labeler from \citeauthor{Peters2018ELMO} (\citeyear{Peters2018ELMO}), achieving an F1 of 84.6\%\footnote{This result nearly reaches the SOTA in \cite{He2018Jointly}.} on English \emph{OntoNotes v5.0} benchmark dataset \cite{pradhan2013towards} for the CoNLL-2012 shared task. At test time, we perform Viterbi decoding to enforce valid spans using BIO constraints. In our implementation, there are 104 labels in total. We use \emph{O} for non-argument words and \emph{Verb} label for predicates.

\begin{table*}
	\centering
	\resizebox{\linewidth}{!}
	{
		\begin{tabular}{lccccccccc}
			\hline
			
			\hline
			\textbf{Method} &  \multicolumn{2}{c}{\textbf{Classification}} &\multicolumn{3}{c}{\textbf{Natural Language Inference}} & \multicolumn{3}{c}{\textbf{Semantic Similarity}} &  \textbf{Score}\\
			
			& CoLA & SST-2 & MNLI & QNLI & RTE  & MRPC  & QQP & STS-B & -\\ 
			&  (mc) & (acc)	& m/mm(acc) & (acc) & (acc)  & (F1) & (F1) & (pc) & -\\
			\hline
			\multicolumn{10}{c}{\emph{Leaderboard (September, 2019)}} \\
			ALBERT & 69.1 & 97.1 &  91.3/91.0 & 99.2 & 89.2  & 93.4 & 74.2  & 92.5  & 89.4\\
			RoBERTa & 67.8 & 96.7 & 90.8/90.2 & 98.9 & 88.2 & 92.1 & 90.2 & 92.2 & 88.5 \\
			XLNET & 67.8 & 96.8 & 90.2/89.8 & 98.6 & 86.3 & 93.0 & 90.3 & 91.6 & 88.4 \\
			\hline
			\multicolumn{10}{c}{\emph{In literature (April, 2019)}} \\
			BiLSTM+ELMo+Attn &36.0 & 90.4 &76.4/76.1&79.9& 56.8 &84.9&64.8  & 75.1 &  70.5\\
			GPT& 45.4 & 91.3  & 82.1/81.4 & 88.1 & 56.0 & 82.3  & 70.3 & 82.0 & 72.8\\
			GPT on STILTs & 47.2 & 93.1   & 80.8/80.6 & 87.2 & 69.1 & 87.7 & 70.1 &85.3  & 76.9\\
			MT-DNN & 61.5 & 95.6   & 86.7/86.0 & - & 75.5 & 90.0 & 72.4 & 88.3  & 82.2\\
			\hdashline
			BERT$_\text{BASE}$  & 52.1 & 93.5  & 84.6/83.4 &  - & 66.4 & 88.9 & 71.2  & 87.1 & 78.3\\
			BERT$_\text{LARGE}$ &60.5 & 94.9 &  86.7/85.9 & 92.7 & 70.1  & 89.3 & 72.1 & 87.6 & 80.5\\
			\hline
			\multicolumn{10}{c}{\emph{Our implementation}} \\
			SemBERT$_\text{BASE}$   & 57.8 &93.5 &84.4/84.0 & 90.9 & 69.3  & 88.2 & 71.8 & 87.3 &  80.9\\
			SemBERT$_\text{LARGE}$ & 62.3 & 94.6 & 87.6/86.3 & 94.6 & 84.5 & 91.2 & 72.8 & 87.8 &  82.9\\
			\hline
			
			\hline
		\end{tabular}
	}
	
	\caption{\label{tab:glue} Results on GLUE benchmark. The block \emph{In literatures} shows the  comparable results from \cite{liu2019multi,radford2018improving} at the time of submitting SemBERT to GLUE (April, 2019).
	}
	
\end{table*}

\subsection{Task-specific Fine-tuning} \label{task-specific}
In Section \ref{sec:bert}, we have described how to obtain the semantics-aware BERT representations. Here, we show how to adapt SemBERT to classification, regression and span-based MRC tasks. We transform the fused contextual semantic and LM representations $h$ to a lower dimension and obtain the prediction distributions. Note that this part is basically the same as the implementation in BERT without any modification, to avoid extra influence and focus on the intrinsic performance of SemBERT. We outline here to keep the completeness of the implementation.

For classification and regression tasks, $h$ is directly passed to a fully connection layer to get the class logits or score, respectively. The training objectives are CrossEntropy for classification tasks and Mean Square Error loss for regression tasks. 

For span-based reading comprehension, $h$ is passed to a fully connection layer to get the start logits $s$ and end logits $e$ of all tokens. The score of a candidate span from position $i$ to position $j$ is defined as  $s_i + e_j$, and the maximum scoring span where $j \geq i$ is used as a prediction\footnote{All the candidate scores are normanized by softmax.}. For prediction, we compare
the score of the pooled first token span: $s_{\tt null} = s_0 + e_0$ to the score of the best non-null span $\hat{s_{i,j}}$ =  $ {\tt max}_{j \geq i} (s_i + e_j)$. We predict a non-null answer when  $\hat{s_{i,j}} > s_{\tt null} + \tau $, where the threshold $\tau$ is selected on the dev set to maximize F1.

\section{Experiments}
\subsection{Setup}
Our implementation is based on the PyTorch implementation of BERT\footnote{\url{https://github.com/huggingface/pytorch-pretrained-BERT}}. We use the pre-trained weights of BERT and follow the same fine-tuning procedure as BERT without any modification, and all the layers are tuned with moderate model size increasing, as the extra SRL embedding volume is less than 15\% of the original encoder size. We set the initial learning rate in \{8e-6, 1e-5, 2e-5, 3e-5\} with warm-up rate of 0.1 and L2 weight decay of 0.01. The batch size is selected in \{16, 24, 32\}. The maximum number of epochs is set in [2, 5] depending on tasks. Texts are tokenized using wordpieces, with maximum length of 384 for SQuAD and 128 or 200 for other tasks. The dimension of SRL embedding is set to 10. The default maximum number of predicate-argument structures $m$ is set to 3.

\subsection{Tasks and Datasets}
Our evaluation is performed on ten NLU benchmark datasets involving natural language inference, machine reading comprehension, semantic similarity and text classification. Some of these tasks are available from the recently released GLUE benchmark \cite{wang2018glue}, which is a collection of nine NLU tasks. We also extend our experiments to two widely-used tasks, SNLI \cite{Bowman2015A} and SQuAD 2.0 \cite{Rajpurkar2018Know} to show the superiority.

\paragraph{Reading Comprehension}
As a widely used MRC benchmark dataset, SQuAD 2.0  \cite{Rajpurkar2018Know} combines the 100,000 questions in SQuAD 1.1 \cite{Rajpurkar2016SQuAD} with over 50,000 new, unanswerable questions that are written adversarially by crowdworkers to look similar to answerable ones. For SQuAD 2.0, systems must not only answer questions when possible, but also abstain from answering when no answer is supported by the paragraph. 

\paragraph{Natural Language Inference}
Natural Language Inference involves reading a pair of sentences and judging the relationship between their meanings, such as entailment, neutral and contradiction. We evaluate on 4 diverse datasets, including Stanford Natural Language Inference (SNLI) \cite{Bowman2015A}, Multi-Genre Natural Language Inference (MNLI) \cite{nangia2017repeval}, Question Natural Language Inference (QNLI) \cite{Rajpurkar2016SQuAD} and Recognizing Textual Entailment (RTE) \cite{bentivogli2009fifth}.

\begin{table} 
{
		\resizebox{\linewidth}{!}
	{
			\begin{tabular}{l c c}
				\hline
				
				\hline
				\textbf{Model} & \textbf{EM} & \textbf{F1}\\
				\hline
				\#1 BERT + DAE + AoA$\dagger$  & 85.9 & 88.6 \\
				\#2 SG-Net$\dagger$ & 85.2 & 87.9\\
				\#3 BERT + NGM + SST$\dagger$ & 85.2 & 87.7\\
				\hline
				U-Net \cite{sun2018u} & 69.2 & 72.6\\
				RMR + ELMo + Verifier \cite{hu2018read+}& 71.7 & 74.2 \\
				\hline
				\multicolumn{3}{c}{\emph{Our implementation}} \\
				BERT$_\text{LARGE}$  & 80.5 & 83.6 \\
				SemBERT$_\text{LARGE}$ & 82.4	& 85.2\\
				SemBERT$^*_\text{LARGE}$ & 84.8	& 87.9\\
				\hline
				
				\hline
			\end{tabular}
		}
		{
			\caption{\label{tab:squad2.0} Exact Match (EM) and F1 scores on SQuAD 2.0  test set for single models. $\dagger$ denotes the top 3 single submissions from the leaderboard at the time of submitting SemBERT (11 April, 2019). Most of the top results from the SQuAD leaderboard do not have public model descriptions available, and it is allowed to use any public data for system training. We therefore further adopt synthetic self training\footnotemark[7] for data augmentation, denoted as SemBERT$^*_\text{LARGE}$.}
		} 
}
\end{table}
\footnotetext[7]{\url{https://nlp.stanford.edu/seminar/details/jdevlin.pdf}}

\paragraph{Semantic Similarity}
Semantic similarity tasks aim to predict whether two sentences are semantically equivalent or not. The challenge lies in recognizing rephrasing of
concepts, understanding negation, and handling syntactic ambiguity. Three datasets are used, including Microsoft Paraphrase corpus (MRPC) \cite{dolan2005automatically}, Quora Question Pairs (QQP) dataset \cite{chen2018quora} and Semantic Textual Similarity benchmark (STS-B) \cite{cer2017semeval}.

\paragraph{Classification}
The Corpus of Linguistic Acceptability (CoLA) \cite{warstadt2018neural} is used to predict whether an English sentence is linguistically acceptable or not. The Stanford Sentiment Treebank (SST-2) \cite{socher2013recursive} provides a dataset for sentiment classification that needs to determine whether the sentiment of a sentence extracted from movie reviews is positive or negative.

\subsection{Results} 
Table \ref{tab:glue} shows results on the GLUE benchmark datasets, showing SemBERT gives substantial gains over BERT and outperforms all the previous state-of-the-art models in literature\footnote{We find that MNLI model can be effectively transferred for RTE and MRPC datasets, thus the models for RTE and MRPC are fine-tuned base on our MNLI model.}. Since SemBERT takes BERT as the backbone with the same evaluation procedure, the gain is entirely owing to newly introduced explicit contextual semantics. Though recent dominant models take advance of multi-tasking, knowledge distillation, transfer learning or ensemble, our single model is lightweight and competitive, even yields better results with simple design and less parameters. Model parameter comparison is shown in Table \ref{tab:glue_com}.  We observe that without multi-task learning like MT-DNN\footnote[8]{Since MT-DNN is a multi-task learning framework with shared parameters on 9 task-specific layers, we count the 340M shared parameters for nine times for fair comparison.}, our model still achieves remarkable results. 

Particularly, we observe substantial improvements on small datasets such as RTE, MRPC, CoLA, which demonstrates involving explicit semantics helps the model work better with small training data, which is important for most NLP tasks as large-scale annotated data is unavailable.

\begin{table}
		\resizebox{\linewidth}{!}
	{
	\begin{tabular}{p{5cm} p{1cm} p{1cm}}
		\hline
		
		\hline
		\textbf{Model} &  \textbf{Dev} & \textbf{Test}   \\ 
		\hline
		\multicolumn{3}{c}{\emph{In literature}} \\
		DRCN \cite{kim2018semantic} &- & 90.1\\
		SJRC \cite{zhang2019explicit} & - & 91.3 \\
		MT-DNN \cite{liu2019multi}$\dagger$ &92.2 & 91.6\\
		\hline
		\multicolumn{3}{c}{\emph{Our implementation}} \\
		BERT$_\text{BASE}$ & 90.8 & 90.7\\
		SemBERT$_\text{BASE}$ & 91.2 & 91.0 \\
		\hdashline
		BERT$_\text{LARGE}$  & 91.3  & 91.1  \\
		SemBERT$_\text{LARGE}$ & 92.0  & 91.6 \\
		\hdashline
		BERT$_\text{WWM}$ & 92.1 & 91.6 \\
		SemBERT$_\text{WWM}$ & 92.2 & 91.9 \\ 
		\hline
		
		\hline
	\end{tabular}
}
	{
		\caption{\label{tab:snli} Accuracy on SNLI dataset. Previous state-of-the-art result is marked by $\dagger$. Both our SemBERT and BERT are single models, fine-tuned based on the pre-trained models.} } 
	
\end{table}

\begin{table} 
	\centering
			\begin{tabular}{l c c c}
				\hline
				
				\hline
				\textbf{Model} &  \textbf{Params} & \textbf{Shared }  &\textbf{Rate}   \\ 
				&  (M)  &  (M) & \\
				\hline
				MT-DNN & 3,060  & 340 & 9.1\\
				BERT on STILTs & 335 & - & 1.0\\
				BERT & 335 & - &  1.0\\
				SemBERT & 340 & - & 1.0\\
				\hline
				
				\hline
			\end{tabular}
		{
			\caption{\label{tab:glue_com} Parameter Comparison on LARGE models. The numbers are from GLUE leaderboard (\url{https://gluebenchmark.com/leaderboard}).}
		}
 \end{table}

\begin{table*}
	\centering
	{
		\begin{tabular}{l c c}
			\hline
			
			\hline
			Question & Baseline  & SemBERT\\
			\hline
			What is a very seldom used unit of mass in the metric system? & The ki &metric slug\\
			What is the lone MLS team that belongs to southern California? &Galaxy  &LA Galaxy\\
			How many people does the Greater Los Angeles Area have?  & 17.5 million&over 17.5 million\\
			
			\hline
			
			\hline
		\end{tabular}
	}
	
	\caption{\label{tab:ans} The comparison of answers from baseline and our model. In these examples, answers from SemBERT are the same as the ground truth.}
\end{table*}

	\begin{table}
		\centering

		{
			\begin{tabular}{l c c c}
				\hline
				
				\hline
				\multirow{2}{*}{\textbf{Model}} &\textbf{SNLI} &\multicolumn{2}{c}{\textbf{SQuAD 2.0}} \\
				& \textbf{Dev}   &  \textbf{EM} & \textbf{F1}\\
				\hline
				BERT$_\text{LARGE}$ & 91.3 & 79.6 & 82.4 \\
				BERT$_\text{LARGE}$+SRL & 91.5 & 80.3 & 83.1 \\ 
				SemBERT$_\text{LARGE}$ &  92.3  & 80.9 & 83.6 \\    
				\hline
			\end{tabular}
		}

		\caption{\label{tab:ablation} Analysis on SNLI and SQuAD 2.0 datasets.}
	\end{table}
Table \ref{tab:squad2.0} shows the results for reading comprehension on SQuAD 2.0 test set\footnote[9]{There is a restriction of submission frequency for online SQuAD 2.0 evaluation, we do not submit our base models.}. SemBERT boosts the strong BERT baseline essentially on both EM and F1. It also outperforms all the published works and achieves comparable performance with a few unpublished models from the leaderboard.

Table \ref{tab:snli} shows SemBERT also achieves a new state-of-the-art on SNLI benchmark and even outperforms all the ensemble models\footnote[10]{\url{https://nlp.stanford.edu/projects/snli/}. As ensemble models are commonly composed of multiple heterogeneous models and resources, we exclude them in our table to save space.} by a large margin.

\section{Analysis}\label{sec:ana}
\subsection{Ablation Study}
To evaluate the contributions of key factors in our method, we perform an ablation study on the SNLI and SQuAD 2.0 dev sets as shown in Table \ref{tab:ablation}. Since SemBERT absorbs contextual semantics in a deep processing way, we wonder if a simple and straightforward way integrating such semantic information may still work, thus we concatenate the SRL embedding with BERT subword embeddings for a direct comparison, where the semantic role labels are copied to the number of subwords for each original word, without CNN and pooling for word-level alignment. From the results, we observe that the concatenation would yield an improvement, verifying that integrating contextual semantics would be quite useful for language understanding. However, SemBERT still outperforms the simple BERT+SRL model just like the latter outperforms the original BERT by a large performance margin, which shows that SemBERT works more effectively for integrating both plain contextual representation and contextual semantics at the same time. 

\begin{table} 
	\centering
		\resizebox{\linewidth}{!}
	{
	\begin{tabular}{l c c c c c}
		\hline
		
		\hline
		Number & 1 & 2 & 3 & 4 & 5   \\ 
		\hline
		Accuracy & 91.49 & 91.36 & 91.57 & 91.29 & 91.42 \\
		\hline
		
		\hline
	\end{tabular}
}
	{
		\caption{\label{fig:m} The influence of the max number of predicate-argument structures $m$.}
	}
\end{table}

\subsection{The influence of the number $m$}
We investigate the influence of the max number of predicate-argument structures $m$ by setting it from 1 to 5. Table \ref{fig:m} shows the result. We observe that the modest number of $m$ would be better.

\subsection{Model Prediction}\label{span_out}
To have an intuitive observation of the predictions of SemBERT, we show a list of prediction examples on SQuAD 2.0 from baseline BERT and SemBERT\footnote[11]{Henceforth, we use the SemBERT* model from Table \ref{tab:squad2.0} as the strong and challenging baseline for ablation.} in Table \ref{tab:ans}. The comparison indicates that our model could extract more semantically accurate answer, yielding more exact match answers while those from the baseline BERT model are often semantically incomplete. This shows that utilizing explicit semantics is potential to guide the model to produce meaningful predictions. 
Intuitively, the advance would attribute to better awareness of semantic role spans, which guides the model to learn the patterns like \emph{who did what to whom} explicitly. 

Through the comparison, we observe SemBERT might benefit from better span segmentation through span-based SRL labeling. We conduct a case study on our best model of SQuAD 2.0, by transforming SRL into segmentation tags to indicate which token is inside or outside the segmented span. The result is 83.69(EM)/87.02(F1), which shows that the segmentation indeed works but marginally beneficial compared with our complete architecture.

It is worth noting that we are motivated to use the SRL signals to help the model to capture the span relationships inside sentence, which results in both sides of semantic label hints and segmentation benefits across semantic role spans to some extent. The segmentation could also be regarded as the awareness of semantics even with better semantic span segmentations. Intuitively, this indicates that our model evolves from BERT subword-level representation to intermediate word-level and final semantic representations.

\subsection{Infulence of Accuracy of SRL}
Our model relies on a semantic role labeler that would influence the overall model performance. To investigate influence of the accuracy of the labeler, we degrade our labeler by randomly turning specific proportion [0, 20\%, 40\%] of labels into random error ones as cascading errors. The F1 scores of SQuAD are respectively [87.93, 87.31, 87.24]. This advantage can be attributed to the concatenation operation of BERT hidden states and SRL representation, in which the lower dimensional SRL representation (even noisy)  would not affect the former one intensely. This result indicates that the LM can not only benefit from high-accuracy labeler but also keep robust against noisy labels.

Besides the wide range of tasks verified in this work, SemBERT could also be easily adapted to other languages. As SRL is a fundamental NLP task, it is convenient to train a labeler for main languages as CoNLL 2009 provides 7 SRL treebanks. For those without available treebanks, unsupervised SRL methods can be effectively applied. For out-of-domain issue, the datasets (GLUE and SQuAD) that we are working on cover quite diverse domains, and experiments show that our method still works.

\section{Conclusion}\label{sec:concl}
This paper proposes a novel semantics-aware BERT network architecture for fine-grained language representation. Experiments on a wide range of NLU tasks including natural language inference, question answering, machine reading comprehension, semantic similarity and text classification show the superiority over the strong baseline BERT. Our model has surpassed all the published works in all of the concerned NLU tasks. This work discloses the effectiveness of semantics-aware BERT in natural language understanding, which demonstrates that explicit contextual semantics can be effectively integrated with state-of-the-art pre-trained language representation for even better performance improvement. Recently, most works focus on heuristically stacking complex mechanisms for performance improvement, instead, we hope to shed some lights on fusing accurate semantic signals for deeper comprehension and inference through a simple but effective method. 

\bibliographystyle{aaai}
\bibliography{sembert}

\end{document}